\documentclass[conference]{IEEEtran}
\usepackage[noadjust]{cite}
\usepackage{amsmath,amssymb,amsfonts}
\usepackage{algorithmic}
\usepackage{graphicx}
\usepackage{textcomp}
\usepackage{xcolor}
\usepackage{fancyhdr}
\usepackage[hyphens]{url}
\usepackage{subcaption}
\usepackage{comment}

\def\BibTeX{{\rm B\kern-.05em{\sc i\kern-.025em b}\kern-.08em
    T\kern-.1667em\lower.7ex\hbox{E}\kern-.125emX}}

\fancypagestyle{firstpage}{
  \fancyhf{}

  \fancyfoot[C]{\thepage}
}  

\title{ Low-Precision Hardware Architectures Meet Recommendation Model Inference at Scale} 
\author{
Zhaoxia (Summer) Deng,
Jongsoo Park,
Ping Tak Peter Tang,
Haixin Liu,
Jie (Amy) Yang,\\
Hector Yuen,
Jianyu Huang,
Daya Khudia, 
Xiaohan Wei,
Ellie Wen,
Dhruv Choudhary,\\
Raghuraman Krishnamoorthi, 
Carole-Jean Wu,
Satish Nadathur,
Changkyu Kim,
Maxim Naumov,\\
Sam Naghshineh, 
Mikhail Smelyanskiy \\
Facebook, 1 Hacker Way, Menlo Park, CA
}

\newcommand{\unComment}[2]{}

\newcommand{\scale}{{\rm scale}}
\newcommand{\zeropoint}{{\rm zeropt}}
\newcommand{\bias}{{\rm bias}}
\newcommand{\clip}{{\rm clip}}
\newcommand{\round}{{\rm rnd}}
\newcommand{\Xmin}{X_{\rm min}}
\newcommand{\Xmax}{X_{\rm max}}

\newcommand{\Imin}{I_{\rm min}}
\newcommand{\Imax}{I_{\rm max}}
\newcommand{\NE}{{\rm NE}}
\newcommand{\NEdiff}{\NE_{\rm diff}}
\newcommand{\quantize}{{\cal Q}}
\newcommand{\dequantize}{{\cal D}}
\newcommand{\Tcomp}{{T_{\rm comp}}}
\newcommand{\Tmem}{{T_{\rm mem}}}
\newcommand{\NI}{\noindent}

\begin{document}
\maketitle
\thispagestyle{firstpage}
\pagestyle{plain}


\begin{abstract}

Tremendous success of machine learning (ML) and the unabated growth in ML model complexity motivated many ML-specific designs in both CPU and accelerator architectures to speed up the model inference. While these architectures are diverse, highly optimized low-precision arithmetic is a component shared by most. Impressive compute throughputs are indeed often exhibited by these architectures on benchmark ML models. Nevertheless, production models such as recommendation systems important to Facebook's personalization services are demanding and complex: These systems must serve billions of users per month responsively with low latency while maintaining high prediction accuracy, notwithstanding computations with many tens of billions parameters per inference. Do these low-precision architectures work well with our production recommendation systems? They do. But not without significant effort. We share in this paper our search strategies to adapt reference recommendation models to low-precision hardware, our optimization of low-precision compute kernels, and the design and development of tool chain so as to maintain our models' accuracy throughout their lifespan during which topic trends and users' interests inevitably evolve. Practicing these low-precision technologies helped us save datacenter capacities while deploying models with up to 5X complexity that would otherwise not be deployed on traditional general-purpose CPUs. We believe these lessons from the trenches promote better co-design between hardware architecture and software engineering and advance the state of the art of ML in industry.

\end{abstract}

\section{Introduction}\label{sec:intro}

Machine learning has experienced revolutionary advances in the past decade. The advancement has led to state-of-the-art production models that are complex, compute intensive and consume large amount of memory resources~\cite{lepikhin2020gshard,yi:sysml2018,zhou2017deep,zhou2019deep,hazelwood2018ml,brown2020language}. For instance, the memory requirement of deep learning recommender systems has grown from few tens of GBs to the terabyte regime~\cite{lui2020understanding,zhao2020distributed}. To fuel the needs of rising application domains, computer architectures are evolving, be it general-purpose CPUs or GPUs with new features~\cite{intel_cooper_lake_bfloat16, frolov2017tensor} or specialized accelerators~\cite{jouppi2017datacenter, aws_inferentia, intel_sph}, targeting deep learning. While these architectural features or architectures are diverse, one major commonality is the support for low-precision arithmetic datatype.

Low-precision arithmetic datatype offers several orthogonal advantages simultaneously. Storing model parameters in low-precision formats greatly increases parameter loading throughput and reduces power consumed on data movement~\cite{han2016deep,hashemi-data2017,sze-cicc2017,brooks-hpca1999,judd-ics2006}. It also increases compute throughput if computation is well optimized for the underlying architecture. For instance, TPUs are customized~\cite{jouppi2017datacenter} with the int8 and fp16 arithmetic to accelerate the inference of neural networks. The latest Intel CPUs added the VNNI feature (Vector Neural Network Instructions)~\cite{intel_vnni} to support neural network acceleration. NVIDIA Tensor Cores support mixed precision of fp32 and fp16 to deliver 7x more flops/s than fp32 on a Tesla V100 GPU~\cite{markidis2018nvidia}. Moreover, int8 is generally supported on many customized accelerators and CPUs, which usually deliver much higher compute throughput than floating point. 

To improve user experience, social network services such as those Facebook provides, leverage deep learning in a wide range of applications~\cite{hazelwood2018ml}. 
Among all the deep learning applications, recent work showed, in 2019, more than 70\% of the total machine learning inference cycles served by Facebook's datacenter fleets are devoted to recommendation use cases~\cite{gupta2020architectural}. 
Thus, performance and energy efficiency improvement of deep learning recommendation models will translate into significant infrastructure capacity saving.

Recommender systems have evolved with increasing model complexities: from simple content and collaborative filtering, to matrix factorization~\cite{Rendle:2010,koren2009matrix}, multi-layer perceptrons and deep learning models~\cite{Weinberger:ICML2009,naumov2019deep,zhou2019deep,dinzhou2018deep,mtwnd,WD:2016,Wang:2018}. Facebook's production recommendation models consist of many tens to a hundred billion parameters to achieve high learning capacity~\cite{DLRM_OCP}. Despite the continuously growing learning capacity, the inference latency of the recommender systems must meet specified SLA thresholds, such as 100ms, to ensure quality user experience~\cite{park2018deep}. 
Low-precision arithmetic is promising to improve the inference efficiency. Nevertheless, deploying low precision in production recommender models at Facebook's scale proves extremely challenging:

\noindent\textbf{Meeting Stringent Accuracy Requirement}: Facebook's recommendation models are typically trained using the full-precision numerical format. Trained models can be converted to a lower precision format for deployment through post-training quantization (PTQ). We do not deploy a low-precision recommendation model that has more than 0.05\% accuracy degradation compared against the original full-precision model~\cite{yangtraining}. This is a much more stringent requirement compared to other vision or language use cases, not to mention that there are many recommendation model variants in production serving for different tasks. Our workflow needs to provide accuracy guarantee for all the production models.
    
\noindent\textbf{Maintaining Consistent Accuracy and Performance during Online Training}:
   The recommendation models are periodically updated via online training to adapt to new data. The online training pipeline includes model training, low-precision model transformation, model publishing and then model serving. 
   In each online training cycle, the workflow that converts a full-precision model to low precision needs not only to produce an accurate and robust model, but also do so quickly to meet the online training cadence.
   
\noindent\textbf{Maximizing Latency-Bounded Throughput Performance}: 
    In order to meet the aforementioned accuracy requirement, a low-precision model in fact often uses mixed precision: exploiting arithmetic types of various precision commensurate with numerical sensitivity of different operators or workload characteristics. Fine-grained performance profiling is needed to guide the low-precision strategies for the different components. Furthermore, to maximize performance gains, we need optimized low-precision kernels tailored for specific hardware architectures. The kernels should support accuracy compensation techniques required by the low-precision operators, and the implementation details should be consistent across the entire system stack. These implementations are challenging by themselves; the numeric validation and debugging support for such implementations on different hardware platforms are even more so.

This paper presents the low-precision techniques, analysis and tool chain we developed to meet the stringent accuracy and performance challenges and to streamline the deployment of Facebook's production recommendation models in low precision on 
existing hardware platforms including CPUs and accelerators. 
We hope the lessons we learned about the good and bad in these architectures are useful and that the methodologies we are sharing are applicable to many low-precision architectures in general. 

The rest of the paper is organized as follows. Section~\ref{low_prec_arch}
introduces the low-precision arithmetic architectures and various fundamental techniques to exploit them; 
Section~\ref{strategies} describes the recommendation model architecture and the important metric any low-precision version must meet. We then present the important strategies we used to deploy our production models in low precision. Section~\ref{acc_tools} describes the supporting tool chain needed to realize these strategies: numerical issues debugging and accuracy maintenance throughout a model's lifespan. Section~\ref{lessons} shares the important lessons we learned in the past couple of years deploying production scale models in low-precision arithmetic. Section~\ref{related} talks about related works on low-precision architectures. Section~\ref{conclusion} recaps the key points and outlines an agenda for future work.

\section{Low-Precision Arithmetic Architectures}\label{low_prec_arch}

Floating-point and integer are the two fundamental data types in computer architectures. Traditional computing often use either the 32 or 64 versions of these type: fp\{32/64\}, int\{32/64\}. Since deep learning is real-number based and the numerical range and precision provided by fp32 are clearly sufficient, that has been the default arithmetic type. More recently, researchers demonstrated that effective deep learning inference~\cite{BFloat16, Jacob_2018_CVPR, jouppi2017datacenter, krishnamoorthi2018quantizing}, and even training~\cite{desa2018highaccuracy, ibm_low_precision_training} can rely on numerics in a more limited range and precision than that given by fp32. 
\begin{figure}
\centering
\vspace{-0.2in}
  \includegraphics[width=0.9\linewidth]{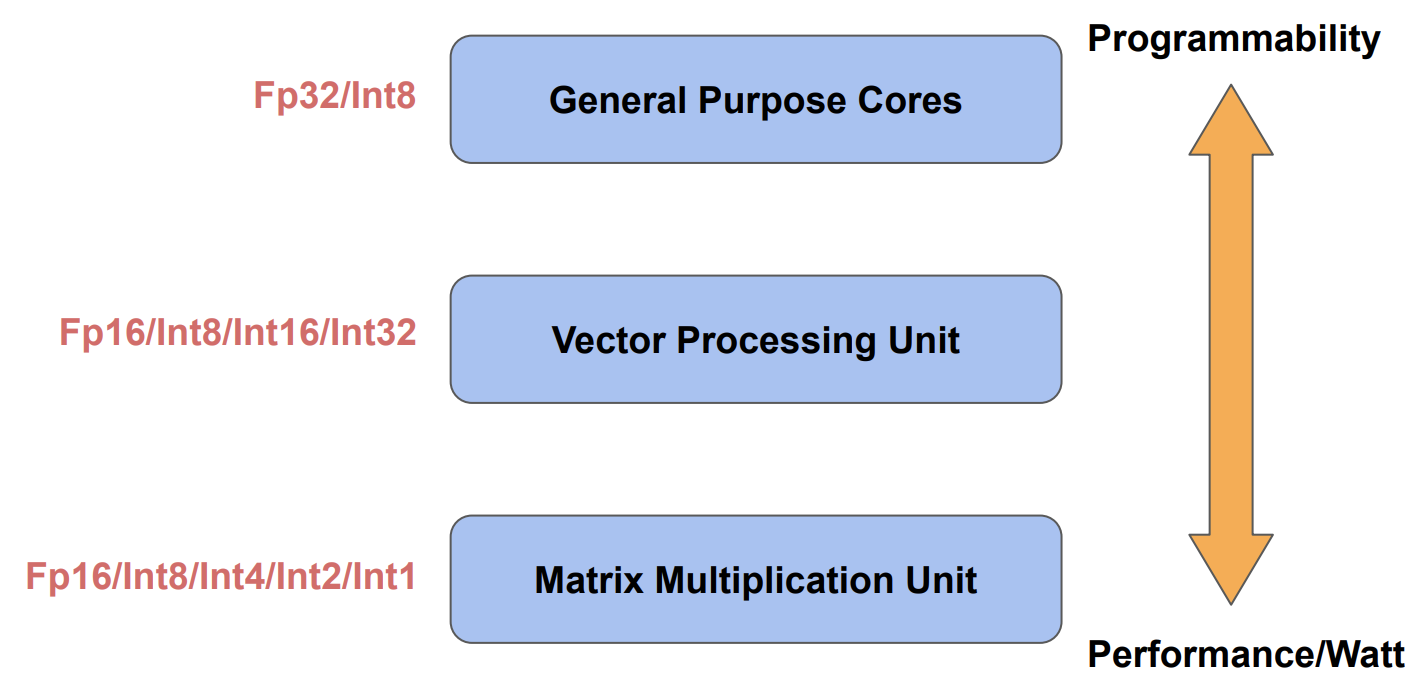}
  \caption{Numerics support on Vendor-X accelerators}
  \label{fig:accelerator_numerics}
\end{figure}
Storage-only low-precision on CPU can be used with appropriate software, thus improving memory access while relying still on fp32 for computation.
Obviously, native arithmetics in the hardware are preferred as they can potentially offer computational savings as well. On traditional Intel general-purpose CPUs,  
low-precision arithmetics support is limited except for int8 and the recent bfloat16~\cite{BFloat16} support on Cooper Lake series~\cite{intel_cooper_lake_bfloat16}. In contrast, customized accelerators have more flexibility to add low-precision arithmetics, and the resulting performance can be significantly higher. Figure~\ref{fig:accelerator_numerics} shows the overall hierarchy of the Vendor-X hardware modules in terms of the compute efficiency and programmability. Notwithstanding the great promises of accelerators, fully exploiting low-precision on CPU is invaluable as they can support continuously emerging new models expeditiously and serve as a stepping stone for those most intensive models to be deployed on customized accelerators. Let us examine these low-precision arithmetic and strategies to benefit from them.


\subsection{Low-precision Floating Point}\label{floating_point}

Low-precision floating-point representations are easy to adopt because of their high programmability. Traditional Intel Broadwell/Skylake CPUs do not support fp16 arithmetic so we can only use fp16 as a storage format. The model can thus be loaded/stored in half precision so that there could be memory size reduction and memory bandwidth savings; but the compute will nevertheless still be done in fp32. In contrast to storage-only support, some custom accelerators support fp16 in storage as well as in computation, offering savings not only in memory footprint and bandwidth, but also boosting compute performance. Indeed, the fp16 arithmetic on accelerators could provide one order of magnitude higher throughput over fp32 on CPUs. A few implementation details are noteworthy:
\begin{itemize}
    \item Down conversion is required when we either convert the full-precision model parameters to fp16, or when we store computed intermediate results (performed in fp32) in memory as fp16. We found it important to saturate values beyond fp16's normal range to the fp16's finite values of the largest magnitude rather than following the default IEEE behaviour that results in $\pm{\rm Inf}$. 
    \item Down conversion of non-zero numbers in fp16's subnormal range needs to be handled consistently: either flushing to zero or saturating at $\pm{\rm fp16\_min}$, the non-zero normal fp16 values of smallest magnitude.
    \item Proper and consistent support for Inf/NaN handling proves important in guarding against abnormal behavior and large accuracy drop of a model.
\end{itemize}

Note in particular that these implementation details may vary on different hardware platforms; maintaining consistency between software implementations with hardware across the entire software stack is important.

Although using fp16 doesn't provide as significant performance wins as short integer formats that will be discussed in later sections (especially the fp16 data storage only on CPUs), it's been a good starting point to migrate the recommendation model inference to the low precision regime and prepare us for the accelerator deployment that relies heavily on low precision numerics to deliver high performance.

\subsection{Short Integer and Quantization}\label{int8_quant}

Alluded to earlier, deep learning intrinsically relies on computing with real numbers. If the representation and computation of these real numbers can use, say, 8-bit integers instead of fp32, the memory saving and performance boost can be superior to the use of fp16. Indeed, on Intel Broadwell/Skylake CPUs, one can expect up to 4x in memory savings and 2x in compute performance boost. Linear quantization is our current method of choice to leverage integer representation and arithmetic to mimic computation of real numbers. 
Two parameters are needed to define linear quantization/dequantization. A common choice is a pair $(\scale,\zeropoint)$ following \cite{gemmlowp} where $\scale$ is a floating-point number and $\zeropoint$ is an integer. Once $(\scale,\zeropoint)$ are chosen, the quantization $X_q = \quantize(X)$ and dequantization $X = \dequantize(X_q)$ operators that map between floating-point values $X$ and integer values $X_q$ are given by Equation~\ref{quant_dequant_scale_zeropoint}:
\begin{align}\label{quant_dequant_scale_zeropoint}
X_q &= \quantize(X) = \clip\left(\round(X/\scale)+\zeropoint,\Imin,\Imax\right), \nonumber \\
X &= \dequantize(X_q) = \scale\times(X_q - \zeropoint).
\end{align}
$[\Imin,\Imax]$ is the range of the target integer type. For example $[\Imin,\Imax]=[-2^{(n-1)},2^{(n-1)}-1]$ for $n$-bit signed integer, and $[0,2^n-1]$ for $n$-bit unsigned integer. 

To obtain the quantization parameters given a floating-point matrix or vector $X$, we first determine its (effective) numeric range $[\Xmin, \Xmax]$. Most straightforwardly, they are indeed the minimum and maximum values in $X$. We may, however, occasionally use a narrower range if for example there are some outliers in $X$. Once $[\Xmin,\Xmax]$ are determined, the parameters $(\scale,\zeropoint)$ are computed below.
\begin{equation}\label{quant_param_compute}
\scale = \frac{\Xmax-\Xmin}{\Imax-\Imin}, \quad
\zeropoint = \round\left(\Imax-\frac{\Xmax}{\scale}\right).
\end{equation}
The quantization of all the fully connected layers is based on Equations~\ref{quant_dequant_scale_zeropoint}and \ref{quant_param_compute} but we use a slight quantization variant when handling embedding tables as shown later.

It is evident that $X - \dequantize(\quantize(X)) \neq 0$ in general and this difference is called the quantization error. There are many important design options to explore in order to reduce the quantization error, such as signed/unsigned integer format, symmetric/asymmetric quantization~\cite{gemmlowp}, the rounding function, etc. Particularly, we discuss the following design options for mitigating the accuracy loss: 

\noindent\textbf{Quantization granularity}: We can use one pair of quantization parameters for the entire matrix or a sub-block of the matrix. For example, per-channel quantization (w.r.t. the output channels in fully connected or convolutional layers) leads to better inference accuracy in practice. The cost, however, is the storage of multiple pairs of the quantization parameters and slightly different arithmetics support on customized accelerators to work with a vector of quantization parameters. 

\noindent\textbf{Quantization range}: In Equation \ref{quant_param_compute}, the $\Xmax$ and $\Xmin$ can either come from the raw data or a refined range eliminating the outliers.
There are many ways to refine the data range. One way is to use the aggregated quantization errors as the guidance. The quantization error can be defined in various forms, e.g. L1-norm, L2-norm or KL divergence. And then we search for the range that minimizes the aggregated quantization errors. Empirically, the L2-norm error minimization algorithm provides the best accuracy for the recommendation models.

\noindent\textbf{Static or dynamic quantization}: The quantization parameters can be chosen either statically or dynamically, based on the history data or the current batch of data for inference. They have pros and cons in terms of both performance and accuracy. For static quantization, we need to insert $Quantize$ and $Dequantize$ operators around the chain of operators to be quantized and pre-calculate the quantization parameters for the inputs and outputs. For dynamic quantization, each of the operators to be quantized will take fp32 inputs, quantize before doing the compute, and then dequantize after the compute to produce fp32 outputs. Dynamic quantization is more flexible in selecting the operators to quantize without runtime overhead from the extra $Quantize$ and $Dequantize$ ops when switching between precisions, and it can adapt to the new data in each online training cycle very well.  However, the drawback is that the overhead of on-the-fly quantization parameter calculation prevents us from using complex quantization algorithms, such as L2-error minimization algorithm, which might provide better accuracy for the model inference. Moreover, without special support, dynamic quantization is more difficult to deploy on hardware accelerators as quantization/dequantization and actual compute may need to be executed on different hardware modules.

\subsection{FBGEMM for Compute Performance}
Low-precision floating-point and integer representations provide clear memory footprint and bandwidth reduction, and careful quantization choices reduce quantization errors. In order to realize the compute performance gain made possible by these reduced precision types, we also need to provide optimized low-precision compute kernels. This leads to our development of FBGEMM~\cite{fbgemm,fbgemm_hpcaml19, fbgemm_blog} library to support the fp16 and int8 operators on Intel CPUs, which has been the backend for the low-precision operators in both Caffe2 and PyTorch. For fp16 operators on CPUs, FBGEMM supports the low-precision conversion and prepacking so that we can reduce the model size and save half of the memory bandwidth when loading the weight parameters. The compute kernels are also generalized to accept the packed weight format. The int8 operators in FBGEMM can also leverage the vector instructions on specific CPU architectures to maximize the compute throughput. Quantization/dequantization operations are fused with the compute kernels when possible. For the customized hardware accelerators, we exploit the kernels that come with the accelerators and provide optimization feedbacks from software-hardware co-design perspectives. We will talk more about accelerator specific optimizations in Section~\ref{accelerators}.

\section{Deployment of Fast and Accurate Recommenders in Low Precision Architecture}\label{strategies}
We aim to substitute recommendation models running in full-precision arithmetic with ones that use low precision for higher serving throughput with little to no degradation of the prediction accuracy. The following section discusses performance and accuracy metrics, overall methodology in utilizing low precision compute, finishing with specific strategies on CPUs and accelerators, respectively. 

\subsection{Performance and accuracy metrics}
For recommendation models, performance is usually measured by the throughput at the p99 latency threshold (e.g. 100ms). Thus, performance optimization techniques such as model partitioning, parallelization, or batching are applied.

The accuracy of recommendation models is evaluated with two kinds of metrics: online and offline. Online accuracy metrics for production models are application dependent, and can only be read after the model is in A/B testing or in production. Consequently, low-precision models must have been evaluated against offline metrics prior being tested online to avoid the potential of significant service quality drop. 
Normalized (cross) entropy (NE)\cite{he2014practical} is an offline accuracy metric that is applicable to various models producing prediction probabilities on binary labeled datasets. Cross entropy is a well accepted loss function for training these models. Normalized entropy is a slight modification, given by Equation~\ref{NE}. 
\begin{equation}\label{NE}
\begin{aligned}
\NE  &= \frac{\sum_{i \in D}{-w_i (y_i log(p_i) + (1-y_i)log(1-p_i))}}{\sum_{i\in D}{-w_i (y_i log(p^*) + (1-y_i)log(1-p^*))}};\\
p* &= \frac{\sum_{i\in D}{w_i y_i}}{\sum_{i \in D}{w_i}};
\end{aligned}
\end{equation}
The numerator is the standard cross entropy where $p_i$ is the prediction probability produced by the model and $y_i = 0$ or $1$ is the label of the data point $\mathbf{d}_i$, $i \in D$, weighted by $w_i$. The denominator is the cross entropy of a naive model that for every data point produces the same probability $p^*$ equal to the fraction of (weighted) positive labeled data. 
NE gives an intuitive indication on how well the model predicts better than the most simple predictions.

The NE metric evaluates how good the predictions are using a specific model, the lower the better. To evaluate the low precision impact, we use the relative NE difference by comparing the NE of a low-precision model with the original fp32 model, as shown in Equation~\ref{NE_diff}. Our goal for the low-precision inference is to maximize the performance given the accuracy constraint. 
One example is to maximize the throughput at 100ms p99 latency while accuracy is within an acceptable threshold (e.g., $\NEdiff \leq 0.05\%$).
\begin{equation}\label{NE_diff}
\NEdiff = \frac{\NE_{\rm lowp} - \NE_{\rm fp32}}{\NE_{\rm fp32}}
\end{equation}

\subsection{Methodology}
The design space of low-precision realizations of a full-precision recommendation model is gargantuan: each of many tens of operators can possibly use low-precision for storage and/or computation in floating point or integer, and the latter with different quantization schemes as described earlier, not to mention the interaction with performance optimizations for each operator. Exhaustive search in the design space is out of the question. Our methodology first heuristically explores low-precision models that meet the $\NEdiff$ accuracy requirement before optimizing the performance of such models on the target hardware platform. Note however that performance optimizations, especially on customized hardware, could affect numerics adversely (e.g. cancellation issues from accumulation order differences) and thus one needs to iterate over the performance and numerics optimizations in general. Given a specific recommendation model and target $\NEdiff$ threshold, Figure~\ref{fig:overall_methodology} depicts the process which we now elaborate:
\begin{figure}
\centering
\vspace{-0.2in}
  \includegraphics[width=0.7\linewidth]{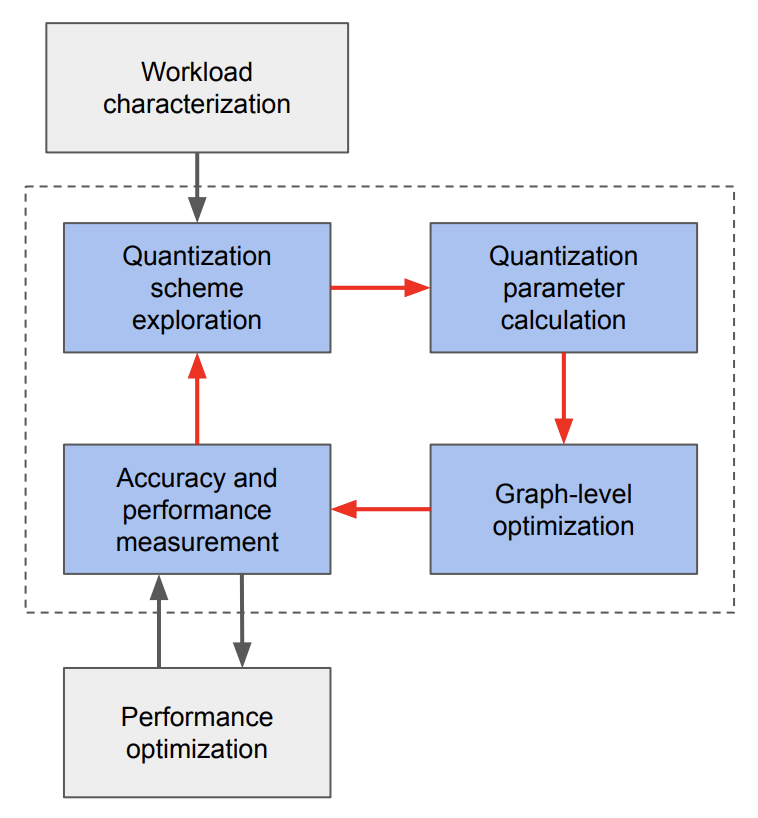}
  \caption{Overall methodology of low-precision optimizations}
  \label{fig:overall_methodology}
\end{figure}

\noindent\textbf{Initial Characterization:} We examine the specific operators employed by the given model and arrive at basic theoretical performance characteristics such as memory intensity or footprint and compute intensity using standard techniques such as roofline analysis~\cite{roofline}\cite{park2018deep}. Together with actual profiling, we identify the top performance bottlenecks. Profiling on the target hardware is easier said than done as more often than not the accelerators are not available or the actual low-precision kernels have yet to be written. Thus, we have developed our internal performance characterization and modeling tool for the accelerator based on the CPU profiling reference and the hardware specs ~\cite{park2018deep}, which allows us to identify such performance bottlenecks early on.

\noindent\textbf{Quantization Scheme:} As int8 offers the best performance improvement opportunities, we aggressively explore the possibility of adopting quantization to all operators and use fp16 (or even fp32) as contingency. Based on the understanding obtained in the previous step about performance bottlenecks, we decide on the precision and specific quantization options if necessary for each of the operators in question. We design the heuristic-based iterative quantization scheme search algorithm and automate the process with the automatic quantization workflow, which we will discuss in more details in Section~\ref{acc_tools}.

\noindent\textbf{Quantization Parameters:} For the int8 quantization, specific quantization parameters are calculated based on the quantization schemes together with histogram of intermediate outputs to determine the effective numerical ranges $[\Xmin,\Xmax]$. Note that an operator that behaves numerically faithful to the underlying low-precision computation needs to be present. It could either be the production quantized operator or some form of efficient emulation. 

\noindent\textbf{Graph Level Optimizations:} Operator fusion is commonly used to improve both accuracy and performance. For example, the $FullyConnected$ operator can be fused with the following $Relu$ operator when possible.  Quantization parameters are subsequently based on $Relu$'s output range rather than that of $FullyConnected$'s, thus better utilizing the low-precision bits. There are also some cases where we can fuse a $\quantize()$ operator that is next to a $\dequantize()$ operator.

\noindent\textbf{Acceptance Test:} Terminate if we have a low-precision production model that satisfies the accuracy constraint. Otherwise, continue improving the quantization scheme based on the feedback from per-layer quantization errors and heuristics to mitigate the errors with advanced quantization options as listed in Section~\ref{int8_quant}. However, if the ratio of the skipped FCs in terms of compute flops exceeds a threshold such that the performance gain from the int8 model diminishes, the iterative quantization scheme search algorithm fails. We need to further investigate on the model and possibly leverage techniques such as quantization-aware training or model co-design to unlock the low-precision techniques.

\noindent\textbf{Performance Optimizations:} This mostly involves kernel optimization on the target hardware platform. Sometimes, the performance optimization involves fundamental changes in the operator implementation, especially on accelerators, which may change the numeric behavior of the operator. At this point, one may return to the early step where different quantization schemes are considered.

Depending on the specific quantization scheme, dynamic or static, fixed quant scheme or adaptive, the optimization cycle can take from seconds to a few hours on a single server-class machine. For instance, static quantization with fixed quant schemes only need a few minutes to collect activation histograms and then calculate the quantization parameters. For static quantization with the auto quantizer, in addition to the histogram collection, the iterative quant scheme search may take $2\sim5$ hours depending on whether we need many iterations to skip FCs in order to meet the accuracy target.
\subsection{Workload characterization for recommendation models}\label{characteristics}
\begin{figure}
\centering
\vspace{-0.3in}
\includegraphics[width=\linewidth]{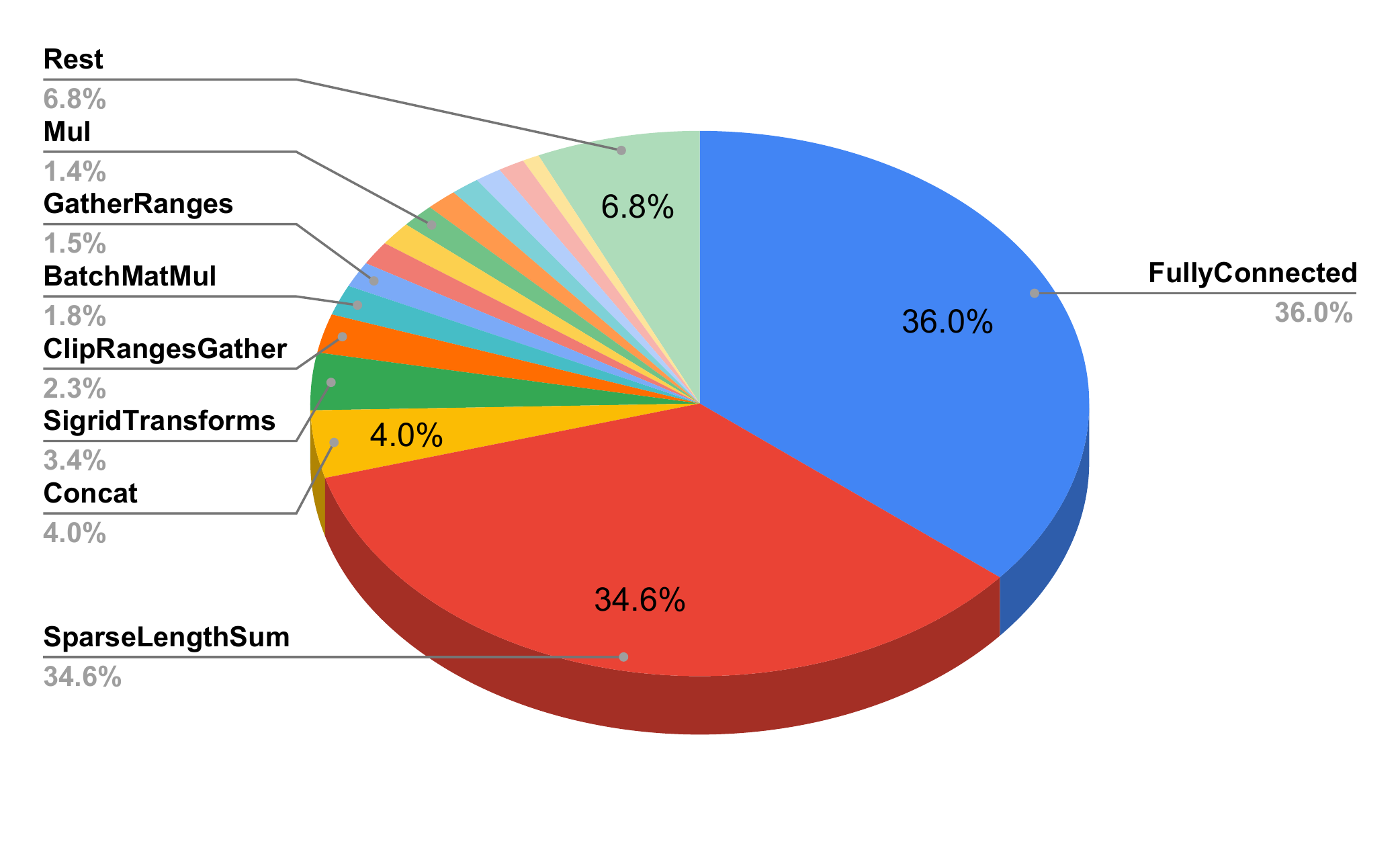}
\vspace*{-0.35in}
\caption{\label{fig:op_latency_breakdown} {Operator latency breakdown for the recommendation models in production}}
\end{figure}

Recent recommendation model size can easily exceed 100GBs, dominated by the hundred-billion-parameter embedding tables. The representative architecture of the recommendation models has been described in~\cite{naumov2019deep},
mainly consisting of embedding table lookups and MLPs so as to operate well on both sparse and dense features. 
Workload characteristics of these components~\cite{park2018deep} differ. Embedding table lookups are memory capacity and bandwidth bound. While MLPs are significantly more compute intensive than embedding table lookups are, they may also be memory bandwidth bound depending on the inference batch size. When no batching is used or the batch size is small, loading the weight parameters from memory could be the main performance bottleneck. Based on our performance roofline model \cite{park2018deep}, batching can shift the MLPs from the memory bound to compute bound, which boosts performance in general. However, the batch size cannot be too large in practice because processing too many entities at once for a single user results in high latency, degrading user experience. 
Figure~\ref{fig:op_latency_breakdown} is one typical profiling snapshot of a particular production recommendation model. It shows the diversity of operators and also underlines the dominance of the $SparseLengthsSum$ and $FullyConnected$ operators, steering our special attention to them.

A suite of synthetic models, representative of production use cases are released as DLRM benchmarks \cite{DLRM_bench} to facilitate the research in external communities. We have provided representative model architecture parameters and performed in-depth performance analysis in previous papers \cite{gupta2020architectural}\cite{gupta-isca2020}. DLRM has been used as the recommendation benchmark for training and inference by the MLCommons community. Specific sizes are adopted based on industry discussions across multiple companies, including Google, Netflix and Alibaba.

\subsection{Low-precision strategies on CPUs}\label{CPU}

\NI\textbf{Int8/Int4 for embedding tables}
We use low precisions for embedding tables to reduce memory footprint rather than economize computation.
Production recommendation models have over a hundred embedding tables, each having anywhere from hundreds to tens of billions parameters. We have explored 8-bit and 4-bit quantization for the embedding tables~\cite{guan2019posttraining}. 


Alluded to previously, quantization and dequantization of embedding tables use a formula slightly different from Equations~\ref{quant_dequant_scale_zeropoint} and~\ref{quant_param_compute}. Specifically,
The quantization and dequantization equations are shown below. 
\begin{align}\label{emb_quant}
X_q &= \quantize(X) = \clip\left(\round((X - \bias)/\scale),\Imin,\Imax\right), \nonumber \\
X &= \dequantize(X_q) = \scale\times X_q + \bias.
\end{align}

\begin{equation}\label{emb_quant_param_compute}
\scale = \frac{\Xmax-\Xmin}{\Imax-\Imin}, \quad
\bias = \Xmin.
\end{equation}
The values $\Imin,\Imax$ correspond to the integer type we choose. For uint4/8, $\Imin=0$ and $\Imax=2^\ell-1$, $\ell = 4$ or 8, respectively.
Note that the ``$\bias$'' parameter here is in floating point. 
Embedding tables are dense by themselves and most consist of a large number of rows. But each inference operation only aggregates (sums up) from each table a few different rows identified by input sparse features, encapsulated by the $SparseLengthsSum$ operator. The single-row output will travel to the interaction component of the model. Currently, the interaction component mostly contains operators that are not $FullyConnected$ or $SparseLengthsSum$ so we convert the output of embedding table lookups to floating point before feeding the features into the interaction component. Using the floating point parameter $\bias$ instead of the integer parameter $\zeropoint$ here saves the compute for dequantization. 

Since each embedding table can have billions of parameters, using a single $(\scale,\bias)$ pair for the entire table
usually leads to higher-than-acceptable quantization error. Thus, in practice, we do rowwise quantization, 
which is to determine one $(\scale,\bias)$ pair per row, stored adjacent to that row for efficient retrieval.

Embeddings quantized with uint8 have been deployed in our production models since a few years ago. It successfully reduced the model size by nearly 50\% without noticeable accuracy degradation in the final event predictions.

More aggressive quantization techniques using 4-bit or even lower bitwidths can potentially bring more capacity savings, but renders 
the resulting models less likely to meet prescribed accuracy requirement. We used different ways to compensate for the accuracy loss in those cases: 
Different algorithms can be used to determine quantization ranges based on L1 or L2 error (Section~\ref{int8_quant}). 
We can also use mixed precisions for the embedding tables. An important observation is that 4-bit quantization results in less quantization errors when used on large embedding tables than on small ones. 
Hence we quantize aggressively with 4-bit the top-N largest embedding tables only.
In this way, we increase the compression ratio while keeping quantization error small. A comprehensive study has been done in \cite{guan2019posttraining} and the solution with mixed precisions has been adopted for the production models. 
Using int4 quantization on the top-half tables in size, which include most of the larger tables, and int8 for the rest 
further reduces the model sizes by 40\%-50\% compared to quantization exclusively with 8-bit. 

Practically, the data is stored as uint8 by packing multiple elements into one byte. Since there is no native support for 4-bit compute on CPUs, we have developed dedicated kernels for the $SparseLengthsSum$ operators in FBGEMM~\cite{fbgemm} to optimize the compute with low-precision inputs and fuse the dequantization step to get the full-precision outputs.


\NI\textbf{Fp16/Int8 for the MLPs}
The MLPs are compute intensive. They are mainly composed of $FullyConnected$ operators in which matrix multiplications dominate inference latency. However, memory bandwidth could also be the bottleneck when the batch size is small and loading the weights from memory is the main overhead. For models running on CPUs, the input sizes of the $FullyConnected$ operators range from 1M to 100M in terms of bytes to read from DRAM. Typical batch sizes range from 1 to 100. To analyze if the operator is compute bound or memory bandwidth bound, we can do some simple calculations as in Equation~\ref{fc_roofline}.

\begin{figure}
\vspace{-0.2in}
  \includegraphics[width=0.95\linewidth]{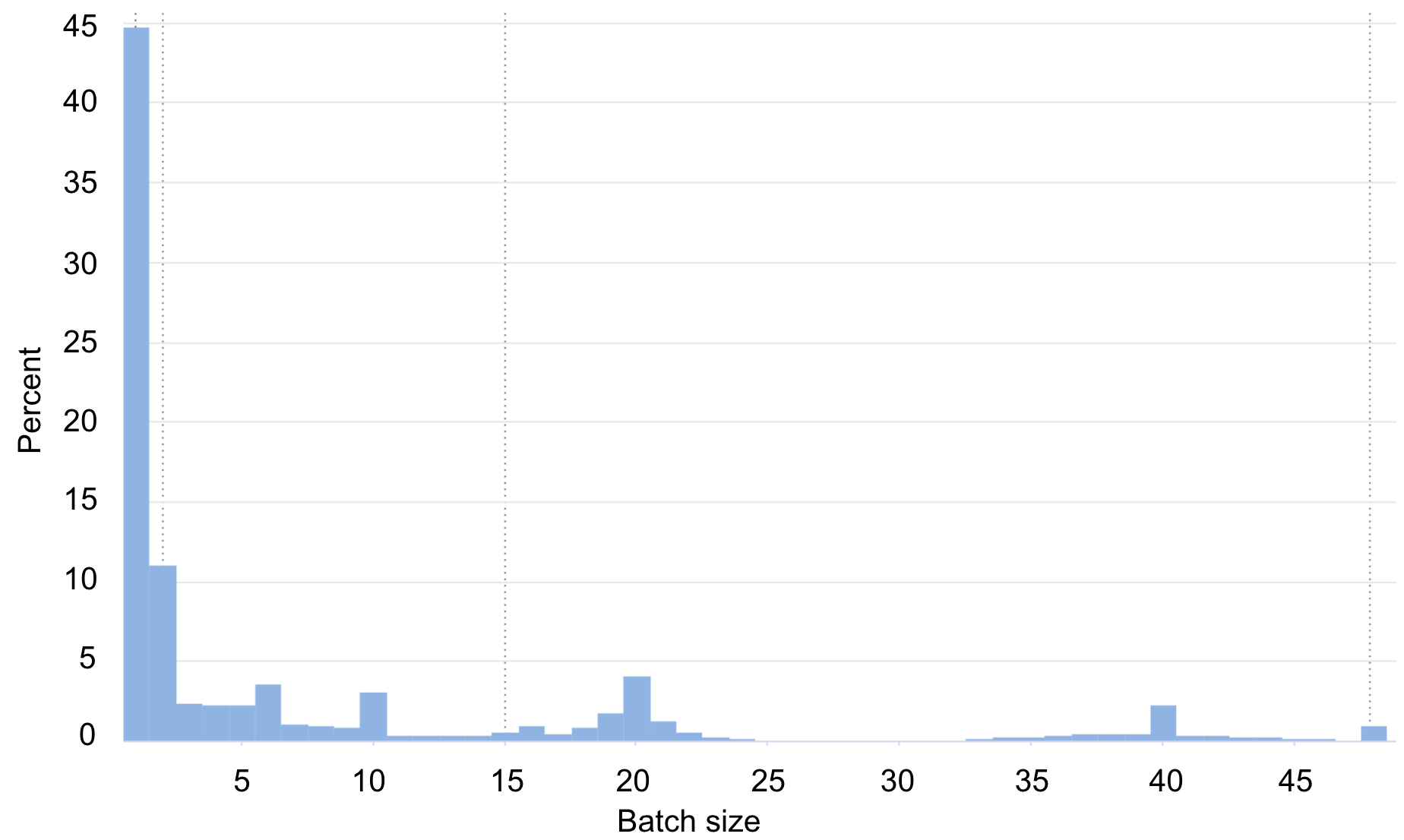}
  \caption{Typical batch sizes for $FullyConnected$ operators}
  \label{fig:fc_batch_size}
\end{figure}
\begin{equation}\label{fc_roofline}
T = \max(\Tcomp, \Tmem) 
    = \max(\frac{2mnk}{F E}, \frac{4kn}{B}),
\end{equation}
where T is the total latency for the operator, $\Tcomp$ is the compute latency, and $\Tmem$ is the memory read latency assuming the parameters are stored in fp32. The operator is to multiply a $m$-by-$n$ input matrix with a $n$-by-$k$ weight matrix. Assume the peak flops of the system is $F$, memory bandwidth is $B$ and compute efficiency is $E$, which is limited by how well instructions map to issue ports. 
The operator becomes memory bandwidth bound when $\Tcomp < \Tmem$, leading to the criteria on batch size $m < 2FE/B$.
For example, on Intel Broadwell CPUs, we have roughly $F=1 Tflops$, $E=90\%$, and $B=70 GB/s$. Hence, the criteria is $m < 2 \times 900/70 \approx 25$. Looking at the typical batch size distribution in Figure~\ref{fig:fc_batch_size}, 44\% of the FCs use batch size 1. And 86\% of operators use a batch size smaller than 25.

Therefore, we want to use low precisions to first mitigate the memory bandwidth overhead, and then improve the compute efficiency after the workload is moved to the compute bound regime.

\NI\textbf{Fp16}: 
CPUs do not support fp16 compute and some do not even support it as a storage format. We therefore handcrafted software to convert and pack/unpack between fp16 and fp32. Loading fp16 weight parameters at runtime is 2X faster than loading fp32 ones for memory bandwidth bound cases. 
Down conversion and prepacking the weights into fp16 only happens once offline and incurs little runtime overhead. The precision for the input and output activations stays in fp32. Our fp16 compute kernels are also released as part of the FBGEMM library~\cite{fbgemm}. Fp16 is one of the highly programmable solutions with little accuracy impact on the model inference, so we have used it for all the recommendation models on CPUs in Facebook datacenters. 

\noindent\textbf{Int8}: Reducing the weights’ precision to int8 can further mitigate the memory bandwidth demand. The 4X bandwidth savings from fp32 to int8 can effectively translate to end-to-end latency reductions for memory bandwidth bound cases. Moreover, as both CPUs and hardware accelerators have native support for int8, we can achieve efficiency wins for compute bound cases as well. To achieve the compute savings, we need to quantize both the weights and activations into int8 and use optimized int8 compute kernels. Weights can be statically quantized and prepacked, but the quantization of the activations needs to be done at runtime. To reduce the runtime overhead, we usually choose the quantization parameters for activations beforehand based on the training data, which is referred to as static quantization. In contrast, dynamic quantization chooses the quantization parameters on the fly, which may incur extra runtime overhead but could adapt to the input data distribution changes during online training. 

Although int8 quantization has the potential to deliver much higher efficiency wins (2-4X on CPUs and an order of magnitude higher on hardware accelerators), delivering sufficient inference accuracy on int8 quantized models is more challenging. 

\begin{figure}
  \includegraphics[width=\linewidth]{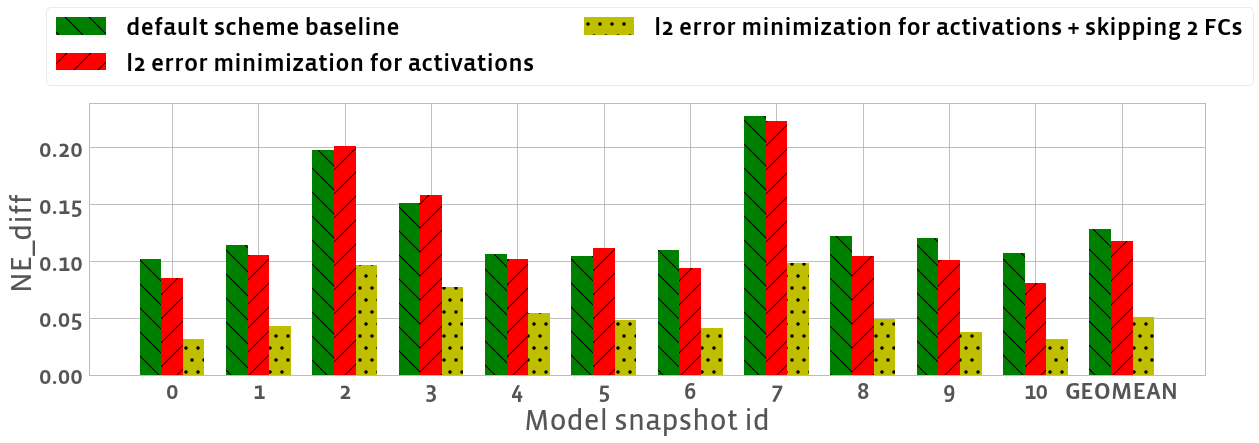}
  \caption{Accuracy impact of different quantization schemes}
  \label{fig:quant_scheme_accuracy}
\end{figure}

As a result, we have explored many different options for the int8 quantization of MLPs. The best quantization algorithm varies for different cases but empirically the L2-error minimization algorithm is more effective for activations and the default min-max algorithm works well for weights. Beyond that, per-channel quantization on weights can further improve the quantization accuracy in cases where the weight data distribution varies across output channels. P99 quantization is effective 
when the raw data's range is large only because of some outliers.
Equalization~\cite{meller2019same} techniques can also help in such cases. Moreover, we use the per layer L2-norm quantization errors to guide the quantization exploration at the model level. We forgo quantizing certain layers, so called ``skipping them'', if not doing so leads to high quantization error. Figure \ref{fig:quant_scheme_accuracy} shows one example of heuristically exploring the quantization schemes on continuous snapshots of a production model and the effectiveness of the schemes in reducing the accuracy gap. The default scheme baseline denotes min-max quantization for weights and activations of FCs.

After choosing the quantization schemes at the operator level, we can further optimize the accuracy and performance at the graph level. Operator fusion is one graph level optimization to improve quantization accuracy. As we have discussed earlier, a FC followed by a Relu can be fused into one operator whose output range is reduced to only the positive part. 
Furthermore, we can optimize the quantized model graph from the performance perspective. $Quantize$ and $Dequantize$ operators are inserted around the int8 operators during the quantization transformation, but these operators will introduce extra performance overhead. So, we try to fuse the quantization operators with the adjacent operators e.g. $Dequantize\to Sigmoid\to Quantize$ when the non-linear activation functions need higher precision. However, to mitigate such overhead, we want to avoid switching numeric precisions in the middle of the model unless necessary. Skipping the last FC in the int8 quantization is common since it does not need extra $Quantize$ or $Dequantize$ operators and the last FC is usually sensitive to the quantization errors.

Unfortunately, there is no easy or clear rule of thumb in identifying the best quantization techniques for all models. 
Thus, 
automating the exploration of appropriate techniques and support for numeric debugging are important;  
Section~\ref{acc_tools} discusses the tools we developed to enable them.

\subsection{Low-precision strategies on inference accelerators}\label{accelerators}

For hardware accelerators, more optimizations can be done based on the specific numerics support on them. We have been collaborating with Vendor-X to deploy their inference accelerators in Facebook datacenters. The Vendor-X accelerators feature large on-chip memory and high memory bandwidth which 
increases inference performance 
of our recommendation models according to the previous analysis in~\ref{characteristics} and~\ref{CPU}.  
\subsubsection{Numerics support on accelerators}
The major numeric formats supported in each compute unit of the accelerator are marked in Figure~\ref{fig:accelerator_numerics}. We try to utilize the int8 and fp16 arithmetics in the matrix core as well as the vector core from where the efficiency gains will mainly come. Fp32 on the general-purpose core is the backup when there are new operators in future models that are not supported in the high-performance units. 

On the vector core where most of the element-wise and non-linear operators are being executed, fp16/int8/int16/int32 are supported to provide high programmability. One limit is that the accumulator for fp16 multiplications is only in fp16, not fp32. This limitation necessitates comprehensive tests to ensure end-to-end accuracy being preserved. For non-linear operators, the vector core supports specialized kernel implementations based on lookup tables and linear or quadratic interpolation. 
On the matrix core, there are systolic MAC arrays with fp16/int8/int4/int2/int1 arithmetics and FMA (fused multiply-add) support. Accumulation into fp32 is supported for fp16 multiplications, which is proved important to the accuracy of MLPs in recommendation models. 
\subsubsection{Model co-design}
The customized accelerators can provide 10X or even higher compute throughput compared to general-purpose CPUs. To fully leverage the high compute power, we need to rethink what kind of models we should run on the accelerators to achieve efficiency wins as well as boost learning accuracies. One general direction is to increase the compute flops for model inference. For example, one of our complex models has 5X more flops than the original model running on CPU. In particular, the $FullyConnected$ operators are wider with more weight parameters and occupy a much higher percentage of the overall inference latency. 
This greatly increased complexity indeed results in a model with much higher prediction accuracy, but one that would not meet our response time constraint if deployed on CPUs. Deployment on accelerators is a solution provided we can leverage the fast low-precision arithmetic (e.g. fp16/int8), as the low-performance fp32 is also inadequate in this aspect.

\subsubsection{Low-precision strategies on Vendor-X accelerators}
According to our internal performance model on the accelerator,
similar to the situation on CPUs,
$FullyConnected$ operators and $SparseLengthsSum$ operators 
also dominate the inference latency so they remain the focus of our numerics and performance optimizations.   

\NI\textbf{Fp16/Int8 for MLPs}: As both int8 and fp16 are supported in the matrix core, we improve the MLP strategies as follows: Start with leveraging int8 quantization on as many $FullyConnected$ operators as possible. However, if an operator causes very high quantization errors, we fall back to fp16 for that operator and insert $Dequantize$ and $Quantize$ operators around it. The fp16 fallback on the accelerator is not the same as the fp16 on CPU because in addition to storing weights in fp16, we exploit the fp16 arithmetic support on the accelerators to achieve a much higher performance. 
To guarantee inference accuracy, we found fp32 accumulation is important for the $FullyConnected$ operators. Otherwise, $\NEdiff$ for some recommendation models can exceed 0.05\%. 
Nevertheless, provided the accuracy constraint is met, exploiting int8 on as many compute intensive operators as possible is important because its throughput is several times higher than fp16. 

\NI\textbf{Fp16 for SparseLengthsSum and other operators}: The embedding tables are stored in the same numeric format as on CPU to save memory capacity and bandwidth. For the reduction operation in $SparseLengthsSum$s, we choose fp16 because compute is not the bottleneck as it is in MLPs and fp16 provides higher accuracy than int8 does. The $SparseLengthsSum$ operators can thus run on the vector core. We have verified that fp16 accumulation in the vector core is sufficient to meet the end-to-end accuracy requirement. Nevertheless, having fp32 accumulation support would have saved significant validation effort. 
We also apply fp16 to a wide range of other operators due to its high programmability, including $BatchMatMul$, $Sigmoid$, $Swish$, $Quantize$, $Dequantize$, etc.
The $BatchMatMul$ operator contributes to the overall inference latency quite noticeably and we could benefit much by quantizing it. This is currently not feasible due to the operator's inputs consisting of dynamic matrices and that the accelerator does not provide a specialized int8 kernel. This lack can in theory be fixed in the future.

\NI\textbf{Graph level optimizations}: Unnecessary switches between numeric precisions result in high overhead because different precisions may need to be processed in different hardware units. In general, we want to apply fp16/int8 to as many operators continuously as possible so that we can maximize the efficiency wins from the high throughput matrix core.

\subsubsection{Numerics validation}
In addition to exploring the best low-precision strategies for our models, a unique challenge of deploying the models on accelerators is to validate that the numeric behaviors of the model inference on the hardware are as expected. 
Validation is particularly important
when the kernel implementations for the hardware from the vendor are not fully transparent. Moreover, the implementations from the vendor could change in different software releases. Although some changes originate from performance optimizations, we need to verify that the numeric behaviors are still correct and consistent after each software release. 

To validate numerics,
we have numeric reference implementations from our side to match with the vendor's implementations deterministically including the accumulation order, and then compare the inference results of our reference implementations with the hardware. Since there are usually graph optimizations and fused operator groups for performance, we also run the model inference end-to-end to match the bit-wise accuracy for the full model. We expect bit-wise accuracy at both the net-level and operator-level, including all the fundamental operators in our recommendation models and commonly used fused operator groups. Bit-wise accuracy matching has been challenging but worth to do to avoid debugging quantization error propagations which could be harder to track. To facilitate the validation process, we have open-sourced our reference implementations and unit tests~\cite{fakelowp} so that the vendor can perform the tests independently after each software release. 

After our comprehensive validation testing against hardware, our reference implementations can serve as a proxy to reflect the device and enable efficient emulations on CPU. Then, we run large-scale evaluations with production traffic and compare the end-to-end accuracy in NE with the CPU baseline. The emulation maps all the CPU operators to reference operators on the accelerator and then run the two models with the same traffic. If the $\NEdiff$ is larger than a threshold, we need to locate the operator that caused the gap and create a repro for the vendor in order to fix the issue in the next software release.

\section{Tooling support for reduced precision optimizations}\label{acc_tools}
With increasing model size and complexity in recommendation models, finding the best quantization strategies for a specific model becomes challenging and often requires much debugging effort during the search. The quantization workflow also needs to handle a variety of production recommendation models aiming at many different tasks. Moreover, recommendation models are usually online trained, and data distribution changes during online training, which can affect the accuracy of specific quantization scheme. Thus, it becomes prohibitive to manually quantize and debug the accuracy issues model by model. To meet these challenges, we developed a suite of tools to help automate the quantization and debugging process. This section first introduces the typical life cycle of models with int8 FCs, and then discusses the tools we have developed to optimize, debug, and monitor the accuracy of int8 recommendation models throughout their life cycle: automatic quantization workflow, numeric debugger, and continuous accuracy monitor. In this section, int8 models denote models with FCs quantized in int8 that need sophisticated flow and tools, while we found that embedding table quantization works well with a fixed scheme.

\subsection{Int8 Recommendation Model Life Cycle}\label{life_cycle}

A typical Int8 recommendation model life cycle involves three major stages: offline experimentation, online experimentation, and model launch.
\subsubsection{Offline Experimentation}
We first do offline experiments, where we quantize the models with various quantization schemes, and then measure offline NE differences of the int8 models from the original fp32 model. This gives us signal on how int8 models perform over the fp32 baseline and helps us choose candidate int8 models for online experiments. This stage uses the {\it automatic quantization workflow} to explore various quantization schemes and the {\it numeric debugger} to deep dive into the per-layer quantization errors when the accuracy cannot meet the requirement.

\subsubsection{Online Experimentation}
In the next stage, 
we put the int8 model obtained above to an A/B test.
In the A/B test, a small portion of real-time production data flows to the int8 model instead of current production model so that we can compare the end-to-end serving quality. 
The fp32 model from where we quantize the int8 model is undoubtedly more accurate than current production model but int8 quantization may introduce accuracy loss. The A/B test allows us to judge if the accuracy gain or performance-accuracy trade-off given by the int8 model warrants its deployment.
This step uses the {\it continuous accuracy monitor} to observe the accuracy of all the int8 model snapshots during online training and uses {\it numeric debugger} for troubleshooting whenever the accuracy drop exceeds the accuracy tolerance. We have encountered more subtle accuracy issues during online experiments than offline because of data distribution shifts. 
Numeric debugger is very useful as it helps
resolve these issues by inspecting the weight/activation distributions, analyzing outlier inputs, and emulating alternative quantization schemes that reduce quantization errors. 

\subsubsection{Model launch}
If the performance and accuracy of the int8 model in the online experiments meet our launch criteria, we roll out the int8 model into production to serve 100\% of real-time data. 
The continuous accuracy monitor is used to guard against model accuracy degradation caused by changes of data distributions, and the numeric debugger will be used when accuracy degradation is considered too large. 

\subsection{Automatic quantization workflow}\label{auto_quant}
\begin{figure}
\centering
  \includegraphics[width=0.9\linewidth]{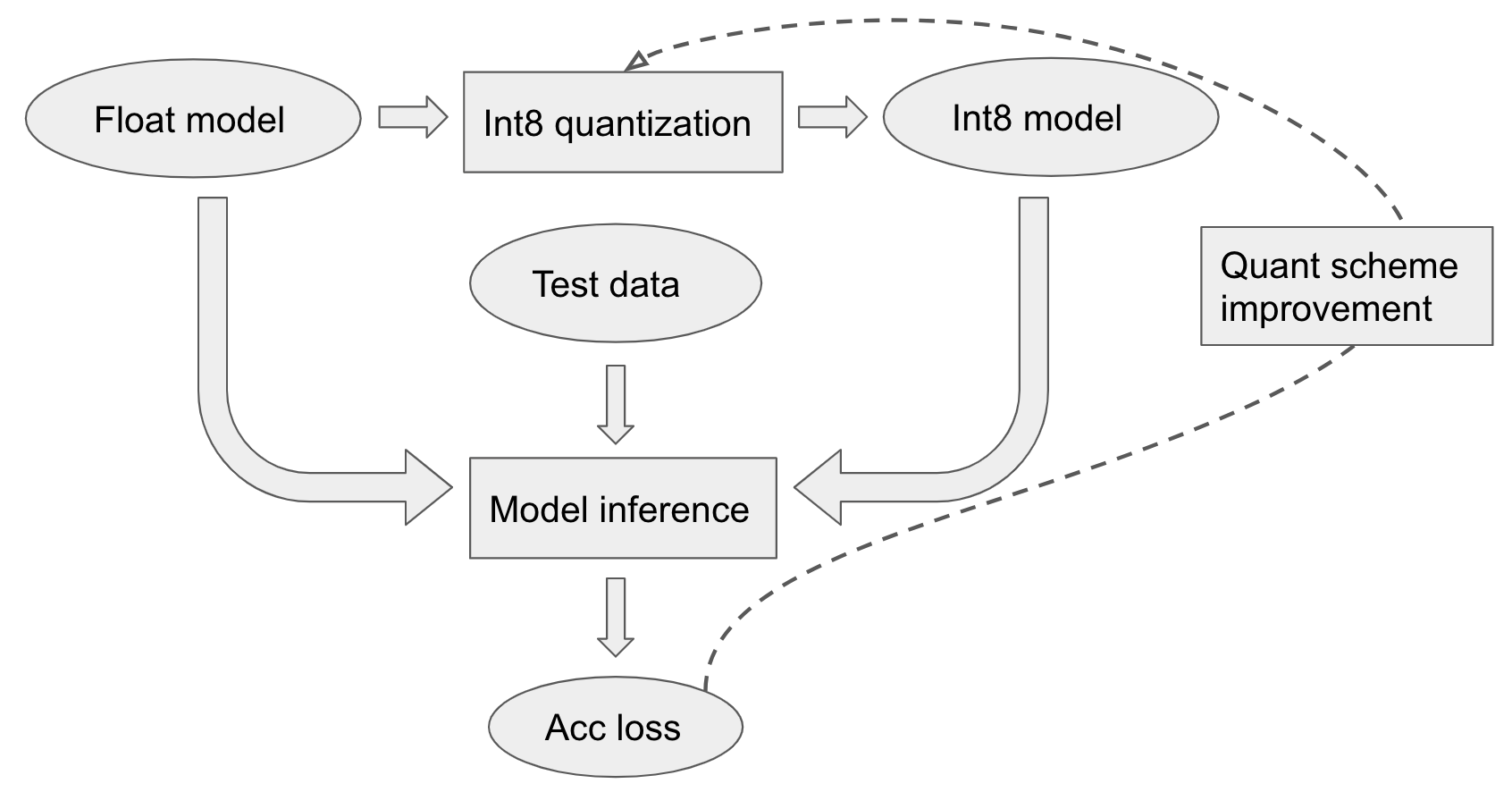}
  \caption{Illustration of the automatic quantization workflow}
  \label{fig:auto_quant_workflow}
\end{figure}

The automatic quantization workflow is an iterative search for a good quantized model among a vast number of candidates. The flow automates data preparation, histogram collection, quantization scheme search, int8 model transformation, accuracy evaluation, per-layer quantization error measurement, and so on. A model candidate's speed and accuracy in a current iteration are assessed and its quantization scheme is further refined if needed. Figure~\ref{fig:auto_quant_workflow} shows this workflow. The search comprises of two main stages as follows.

\subsubsection{Global Quantization Scheme Search}
During this stage, weights and activations in all the layers use the same quantization scheme any given time, hence the name global. This greatly reduces the search space by avoiding exponential combinations of varying schemes at each layer. An example global quantization scheme is L2 error minimization algorithm with asymmetric quantization for each layer and skipping the last layer. We enumerate combinations and measure the accuracy as follows. For each global quantization scheme, the model is transformed into an int8 model and evaluated on a small evaluation dataset. The scheme that provides the highest accuracy will be selected as the best global quantization scheme. If the accuracy loss is acceptable, then the int8 model is saved and the quantization process stops. Otherwise, we proceed to the second stage.
\subsubsection{Iterative Quantization Scheme Search}
We identify the layers with the highest per-layer quantization errors. Then, we apply techniques that can potentially improve the accuracy such as channel-wise quantization on weights and p99 on input/output activations for a layer. If the accuracy is still not good enough, we skip quantizing this layer. The stopping criteria of the iterative search is based on the accuracy loss and the 
percentage of compute flops we need to skip. 
When percentage is high, the performance gain from the int8 quantization becomes marginal hence we stop the search.

Two small datasets are used to make the search time manageable: a calibration dataset, sampled from the training data, is used for histogram collection and quantization parameter calculation; an evaluation dataset, sampled from the offline evaluation data, is used for accuracy measurement. After getting an int8 model with satisfactory accuracy on the small evaluation dataset, we then check the accuracy on full evaluation dataset. If accuracy gap happens between the two evaluation datasets, we draw more samples in the small datasets and repeat the quantization scheme search. 
 

\subsection{Numeric debugger}

\begin{figure}
\centering
  \includegraphics[width=0.6\linewidth]{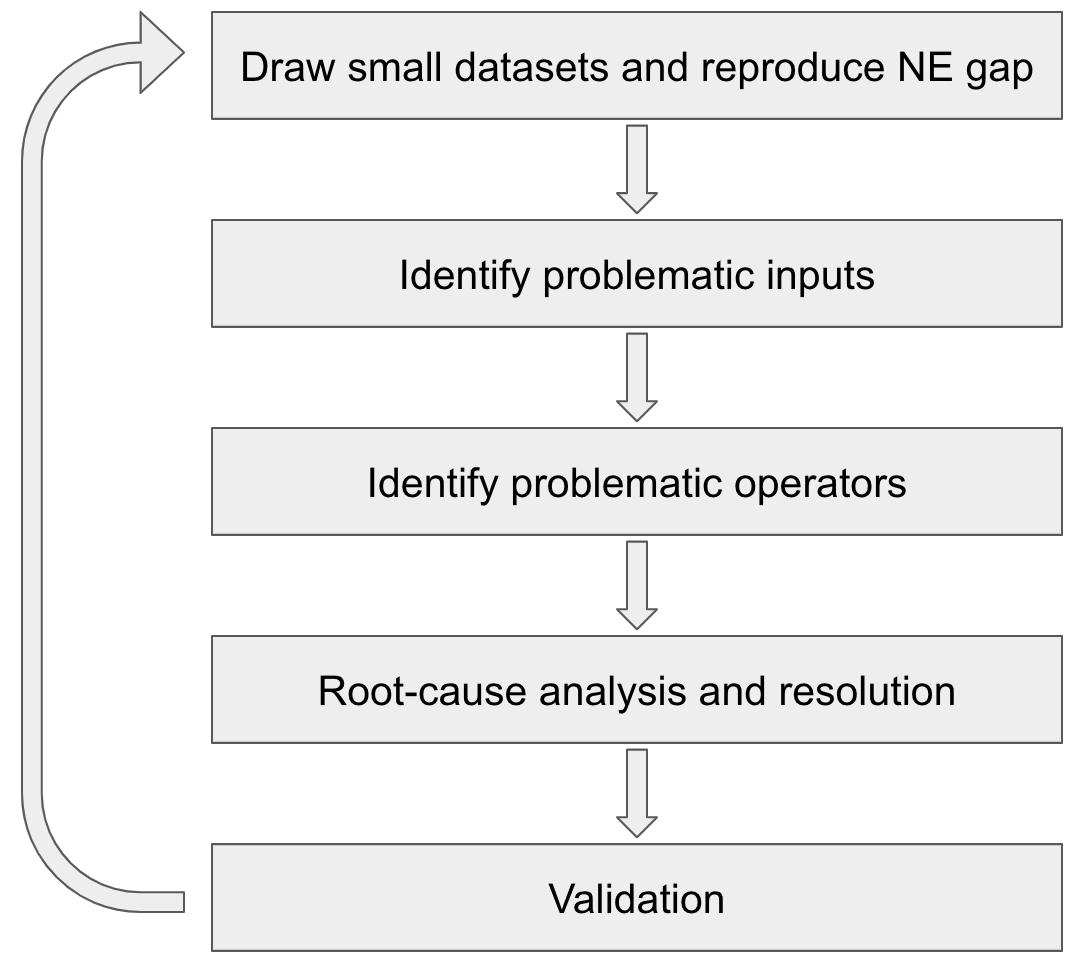}
  \caption{Methodology of the numeric debugger}
  \label{fig:numeric_debugger}
\end{figure}

Debugging numerical accuracy issues must be carried out quickly as there are many production models, all updated frequently. The sizes of these models and the related dataset pose a great challenge as merely loading them can take hours. Another challenge is that we aim to identify the root cause of accuracy drop as oppose to finding superficial temporary fix. This will help us accumulate effective measures for accuracy restoration over time. The debugging methodology we adopted is shown in Figure \ref{fig:numeric_debugger} and discussed in detail below.

\subsubsection{Reproduce $\NEdiff$ on a small dataset}
When the accuracy drop is beyond the prescribed threshold, we first try to reproduce the results with a small dataset, which we can save locally to quickly iterate on debugging in a deterministic way. 
\subsubsection{Identify the inputs that correspond to large accuracy drops}
We further narrow down by sorting the small dataset based on per sample error. 
We calculate the cross entropy error for each sample based on the predictions from the low-precision and original models. The samples are then sorted by the cross entropy errors in descending order. The top samples contribute the most to the accuracy loss in normalized entropy (NE) as the ${NE_{diff}}$ is basically a summation of the cross entropy errors of all samples divided by a normalizer.
With only a few samples, it is much easier to debug and understand the accuracy loss. 
\subsubsection{Identify the operators that are responsible for accuracy drop}
We narrow down to a few operators based on their quantization error when running with the problematic input. 
In this run, a shadow fp32 operator is executed for each low-precision operator with the same input. The difference in outputs is calculated as the quantization error in this layer. The quantization error is used to sort all the operators in order to locate the top problematic operators, which are often responsible for the global accuracy drop.
\subsubsection{Root cause analysis and resolution}
After we reduce the problem size and narrow down to a few inputs and operators, we try to identify the root cause by profiling the distribution of input, output, and weight values of the problematic operators. Based on the profile, we may need to add advanced options in the quantization scheme search space, debug the kernel implementations, validate the kernels on target hardware, etc. in order to fix the accuracy problem. If still not successful, we need to incorporate other techniques such as quantization aware training~\cite{krishnamoorthi2018quantizing, jain2020TQTmlsys, fan2020training}, equalization~\cite{meller2019same}, or model co-design.
\subsubsection{Validation}
We validate the root cause and fix on the saved local dataset, and then on the full evaluation dataset if the accuracy is good. If there is an accuracy gap between the local and global evaluation datasets, we increase the number of samples in the local dataset and repeat the previous quantization scheme search until the accuracy is good on the full evaluation dataset. 

Our numeric debugger enables the above methodology by providing a numeric analysis toolkit and supporting utilities such as dataset extraction and embedding table shrinkage. It is open sourced and also included in PyTorch 1.6 release \cite{pytorch_numeric_suite}.

\subsection{Continuous accuracy monitor}

Our continuous accuracy monitor ensures that the quantization scheme developed for a particular model maintains good accuracy for all online retrained snapshots of the latter.
The tool quantizes every snapshot and evaluates the fp32 and int8 models side by side, in order to proactively identify accuracy issues 
to avoid prolonged degradation of service quality.
The evaluation results are logged to a Scuba \cite{scuba} table so that we can visualize the time series of the accuracy metrics and identify when the accuracy starts to degrade. To diagnose the accuracy degradation, we need the numeric debugger together with the information from the continuous accuracy monitor to locate the problematic model and input data.

In addition, we also support the int8 emulation for Vendor-X accelerators in this workflow to compare the two model runs using the accelerator emulation kernels and CPU kernels side by side. The statistics on the $\NEdiff$ and per-layer quantization errors can 
guide the numeric validation process and help locate bugs in the kernel implementations.

\section{Lessons learned}\label{lessons}

We summarize here the important lessons on low-precision architectures and the hardware-software co-design process we learned. These lessons are accumulated during our long journey in deploying production models in conjunction with hardware vendors refining their platforms.

\begin{itemize}
    \item  \emph{Enhanced Integer Quantization Support:}
    Computation with int8/int4 provides significant performance gains but we rely on the various quantization techniques to meet our accuracy requirements. Hardware support for these techniques such as asymmetric quantization, fine-grain per-channel quantization and dynamic quantization are important, but not all supported at present.
    \item \emph{Balanced Floating-Point and Integer Performance:} 
    We prefer good/balanced performance on low-precision floating-point compute such as fp16 or bfloat16 to resources devoted to extreme-low-precision integer performance at the 1 or 2-bit realm. Low-precision floating-point has been the easiest fallback for network layers of high numerical sensitivity, allowing us to meet our accuracy constraints.
    \item \emph{High-Capacity and Bandwidth Memory:} 
    We foresee the need for large-input-dimension $FullyConnected$ operators in MLPs which help mitigate the effect of quantization errors. Large on-chip and/or high-bandwidth memory will greatly facilitate this.
    \item \emph{Accurate Non-Linear Functions:}
    We see high sensitivities to quantization errors in non-linear activation functions such as Sigmoid, Swish, etc. when their input ranges are not covered sufficiently. Enough coverage is important despite the needed slight increase in LUT size and the associated increase of real estate and power consumption.
    \item \emph{More Optimized Low-Precision Kernels:} 
    Some important and compute heavy operators such as $BatchMatMul$ in existing and emerging models are not supported at the hardware level.
    \item \emph{Co-Design Support:} Open-source numeric reference implementations from vendors can greatly save engineering effort on the numeric validation of the kernel implementations for the accelerators.
    \item \emph{Beyond Hardware and the Model:} The tool chains that automate quantization and facilitate debugging is important to the success of deploying low-precision models. They are a major undertaking: any project planing on model deployment need to take that into account. 
\end{itemize}


\section{Related work}\label{related}

Increasingly complex learning models spur great interest in academia and industry on exploiting low-precision arithmetic to boost inference efficiency. Most of the many academic works on quantization focus on computer vision~\cite{krishnamoorthi2018quantizing, han2016deep, meller2019same, jain2020TQTmlsys, wang2019haq, wu2020integer, jacob2017quantization} or natural language processing models~\cite{zafrir2019q8bert, zadeh2020gobo, bhandare2019efficient, wu2020integer}. There are several open-source libraries from industry that support low-precision numerics for model inference. Notably, Google open sourced its gemmlowp~\cite{gemmlowp} library for 8-bit matrix multiplications. NVIDIA released its TensorRT~\cite{tensorrt} library which supports both fp16 and int8 optimizations for GPU inference.

In the meantime, ubiquitous deep-learning applications have also motivated novel deep-learning friendly hardware designs. New features are added to traditional hardware platforms. For example, Intel added bfloat16~\cite{intel_cooper_lake_bfloat16} as well as VNNI (Vector Neural Network Instructions)~\cite{intel_vnni} in their latest CPU series. NVIDIA Tensor Core supports both fp16 and int8 to achieve higher inference throughput~\cite{markidis2018nvidia}. A variety of AI accelerators~\cite{jouppi2017datacenter, aws_inferentia} with high compute throughput and power efficiency are designed to speed up DL models. 


On recommendation models in particular, DLRM \cite{naumov2019deep} is representative of Facebook's deep learning recommendation models. It has also been adopted by the industry-wide MLPerf benchmark suites~\cite{mlperf-ieeemicro,wu2020developing,mlperf-inference,mlperf-training}. Recent works~\cite{park2018deep, gupta2020architectural,hsia2020crossstack,gupta-isca2020} presented in-depth system performance characterization studies and shared the architectural implications for additional optimization opportunities. Our work here demonstrates a hardware-software co-design in action on deploying production-scale recommender systems to low-precision hardware architectures. We also make a number of recommendation-model-specific suggestions on hardware architectures based on our work ``in the trenches." 

\section{Conclusion}\label{conclusion}

This paper shared the techniques and software tools we developed for deploying production recommendation models in Facebook datacenters to exploit low-precision arithmetic of both CPUs and accelerators. We learned a number of important lessons and made some concrete suggestions to hardware architects designing for large-scale recommendation models and to software engineers deploying these models. 

The work described here can be classified as post-training quantization to low precision. We are exploring several other threads. One is quantization aware training (QAT), which tries to obtain through the training process an easy-to-quantize or already-quantized model (cf. \cite{choi2018PACT,jain2020TQTmlsys}). Model pruning is a technique orthogonal to quantization, but a pruned model yields the same resource saving benefits a quantized model gives. Along this line, knowledge distillation from a complex teacher model to a pruned student model may maintain the needed accuracy produced by the latter. Agile sparsity adaptation techniques are also needed for pruning to be effective as data distribution can change significantly during online training~\cite{ye2020adaptive}.

From the hardware-software co-design perspective, we cannot over emphasize the benefits of both parties open sourcing~\cite{leiserson2020there} important building blocks of their design and code: making these accessible allows continuous validation of hardware functionality and numerical correctness, greatly shortening the entire development cycle. Co-design does not end when a hardware platform is fully designed or made available: as new software algorithm will be developed to exploit new architecture, whose follow-on version in turn will adapt to the more influential ones of these new algorithms.



\bibliographystyle{IEEEtranS}
\bibliography{refs}

\end{document}